\pdfoutput=1

\documentclass[11pt]{article}

\usepackage[final]{acl}

\usepackage{amsmath}
\usepackage{amssymb}
\usepackage{times}
\usepackage{latexsym}
\usepackage{kotex}
\usepackage{booktabs}
\usepackage{multirow}
\usepackage{lscape}
\usepackage{booktabs}
\usepackage{array}
\usepackage{longtable}
\usepackage{color}
\usepackage{tabularray}

\usepackage{tikz}
\usetikzlibrary{arrows.meta}  

\usepackage[T1]{fontenc}

\usepackage[utf8]{inputenc}

\usepackage{microtype}

\usepackage{inconsolata}

\usepackage{graphicx}
\DeclareMathOperator*{\argmin}{arg\,min}

\title{Semantic Aware Linear Transfer by Recycling \\ Pre-trained Language Models for Cross-lingual Transfer}

\author{Seungyoon Lee, Seongtae Hong, Hyeonseok Moon, Heuiseok Lim\thanks{Corresponding authors} \\
  Korea University, Republic of Korea \\
  \texttt{\{dltmddbs100, ghdchlwls123, glee889, limhseok\}@korea.ac.kr} \\}

\begin{document}
\maketitle
\begin{abstract}
Large Language Models~(LLMs) increasingly incorporate multilingual capabilities, fueling the demand to transfer them into target language-specific models. However, most approaches, which blend the source model’s embedding by replacing the source vocabulary with the target language-specific vocabulary, may constrain expressive capacity in the target language since the source model is predominantly trained on English data. In this paper, we propose \textbf{S}emantic \textbf{A}ware \textbf{L}inear \textbf{T}ransfer~(\textbf{SALT}), a novel cross-lingual transfer technique that recycles embeddings from target language Pre-trained Language Models~(PLMs) to transmit the deep representational strengths of PLM-derived embedding to LLMs. SALT derives unique regression lines based on the similarity in the overlap of the source and target vocabularies, to handle each non-overlapping token's embedding space. Our extensive experiments show that SALT significantly outperforms other transfer methods and achieves lower loss with accelerating faster convergence during language adaptation. Notably, SALT obtains remarkable performance in cross-lingual understanding setups compared to other methods. Furthermore, we highlight the scalable use of PLMs to enhance the functionality of contemporary LLMs by conducting experiments with varying architectures.
\end{abstract}

\section{Introduction}
As Large Language Models~(LLMs) continue to demonstrate remarkable performance across various knowledge-based benchmarks and sub-tasks, the demand for robust linguistic capabilities in multilingual or language-specific contexts has surged \cite{wei2022emergent,wei2022chain,peng2023instruction,taori2023alpaca,zhou2024lima}. However, most multilingual-considered LLMs show limited practicality in target languages due to their English-centric training and reliance on common vocabulary designed to incorporate a wide range of languages~\cite{puttaparthi2023comprehensive,le2023bloom,lai-etal-2023-chatgpt,zhao2024llama,gemmateam2024gemma,dubey2024llama}. This drawback also correlates with the large portion of their parameters, which are the embeddings of irrelevant language tokens to target language.

To address this challenge, a promising line of studies has focused on cross-lingual transfer~\cite{artetxe-etal-2020-cross,tran2020english,gee2022fast,dobler2023focus,remy2024trans,liu2024ofa,mundra2024empirical,ye-etal-2024-mosecrot}. These efforts include replacing a source vocabulary with a target language vocabulary, simply initializing embeddings with newly manipulated embeddings stem from source model~\cite{minixhofer2022wechsel,dobler2023focus,liu2024ofa,mundra2024empirical}. 

However, a newly initialized embedding that relies solely on the source model inevitably experiences limited expressiveness in the target language due to the immature learning process of the source model, which does not mainly concentrate on the target language. Moreover, most approaches adopt small encoder architectures such as BERT-like models~\cite{devlin-etal-2019-bert,liu2019roberta,conneau-etal-2020-unsupervised} as a source model, leaving their applicability on contemporary decoder-based LLMs unexplored~\cite{gogoulou2022cross,gee2022fast,zeng2023greenplm,dobler2023focus,ye-etal-2024-mosecrot}.

In this work, we propose \textbf{S}emantic \textbf{A}ware \textbf{L}inear \textbf{T}ransfer~(\textbf{SALT}), a novel cross-lingual transfer method that leverages the rich representational power of target language embeddings in traditional Pre-trained Language Models~(PLMs)\footnote{In this work, we define PLMs to small-scale language models with only millions of parameters, developed prior to the advent of LLMs.} to convert English-centric LLMs into target language–specialized LLMs. To facilitate the transfer, we identify the semantically closest tokens for each non-shared token from a shared vocabulary space. Based on these semantically similar tokens, we fit a unique linear regression to transfer the embeddings from the PLM’s space to the LLM’s space. This procedure enables non-shared vocabulary to be transferred to LLM space while retaining the semantic representation embedded in PLM embedding.

In our experiments, we focus on investigating the potential of transferred embeddings from PLMs in three aspects: (i) whether the transferred embeddings can be well-aligned with inherent knowledge in the source model, (ii) the influence on better initialization to the target language and convergence during continual pre-training, and (iii) the ability of understanding between target language and mainstream language in cross-lingual environments.

We empirically demonstrate that embeddings from target language PLMs can be more helpful for cross-lingual transfer in LLMs than existing methods. SALT not only surpasses various strong baselines in downstream tasks but also provides a better initialization for target language adaptation with faster convergence during continual pre-training. Notably, we find that SALT preserves English capability more effectively while maintaining alignment between English and the target language in a cross-lingual setup. Furthermore, by conducting additional study on various PLM architectures~(Encoder, Decoder, and Encoder-Decoder), we discover that SALT can be extended to PLMs with various architectures and can even be a valid strategy for basic models such as BERT~\cite{devlin-etal-2019-bert}. Our contributions are as follows:
\begin{itemize}
    \item We propose SALT, a novel embedding transfer method to effectively project the deep representation capabilities of PLM embeddings onto LLMs based on semantic information.
    \item We empirically verify that SALT has superiority in cross-lingual environments as well as downstream tasks and language modeling for the target language.
    \item We further show the versatility of SALT in experiments with various types of PLMs.
    \item By recycling previously prominent PLMs as target language embeddings, we demonstrate the potential scalability and advantage of PLMs, and propose a new approach to leverage PLMs in the era of LLMs.
\end{itemize}

\section{Related Work}
Cross-lingual transfer to a target language includes approaches that manipulate the vocabulary or initialize embedding for the new target vocabulary. In vocabulary manipulation, most research has considered adding target language relevant vocabulary~\cite{wang2020extending,chau-etal-2020-parsing,cui2023efficient,larcher2023cabrita,fujii2024continual,mundra2024empirical,yamaguchi2024vocabulary,zhao2024llama}. Vocabulary expansion is widely adopted for developing language-specific models from a source model~\cite{balachandran2023tamil,cui2023efficient,fujii2024continual}. This line of work requires access to a large amount of target language data. In light of this, \citet{yamaguchi2024can} delve into cross-lingual vocabulary expansion in low-resource settings across initialization approaches and training strategies. Also, \citet{zhao2024llama} investigate training scales required for vocab extension and the influence of transfer on capabilities of language generation and following instructions.

\begin{figure*}[hbt!]
\centering 
\includegraphics[width=\linewidth]{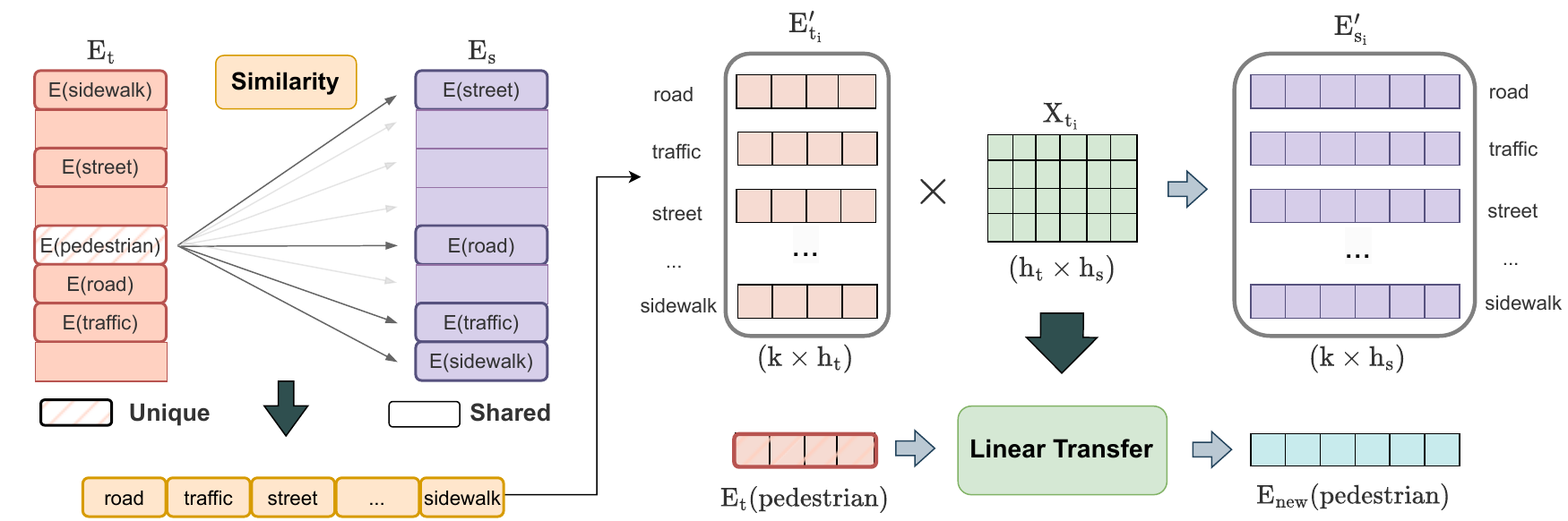}
\caption{Summary of SALT. By using the paired embeddings of semantically similar tokens for each non-shared token, we create a unique least square matrix $X_{t_i}$ to transfer from PLM to LLM. $k$ denotes the number of selected nearest tokens among overlapping tokens for linear transfer, and $h_t$ and $h_s$ refer to the hidden dimensions of the target and source models.}
\label{fig:salt} 
\end{figure*}

On the other hand, given that overall cost increase derived from extended vocabulary, several cross-lingual transfer studies have pursued replacing the original vocabulary with a new target vocabulary~\cite{zeng2023greenplm,ostendorff2023efficient,dobler2023focus,ye-etal-2024-mosecrot,liu2024ofa,remy2024trans}. They focus on initializing new target embeddings by manipulating source embeddings according to semantic similarities between vocabularies. For instance, \citet{gee2022fast} compresses the source model’s vocabulary using an averaging-based technique to accommodate domain-specific terms. 

More recently, lines of work use source embedding and well-aligned external word embeddings to initialize new subword embeddings for a target-specialized vocabulary~\cite{minixhofer2022wechsel,ostendorff2023efficient,ye-etal-2024-mosecrot}. Notably, FOCUS~\cite{dobler2023focus} computes a weighted mean from a multilingual language model, guided by token similarities from fastText~\cite{bojanowski2017enriching}, to obtain new embeddings. Similarly, OFA~\cite{liu2024ofa} adopts a comparable strategy to build multilingual models and proposes a factorization–based dimension reduction embedding transfer method for efficiency. As a hybrid method, \citet{remy2024trans} combines the source model’s embedding under a statistical translation scheme across languages.

However, initializing embedding relying on the source model’s embedding inherently lacks crucial information specialized for the target language since the source embedding does not prioritize target language adaptation. Given this concern, we adopt embeddings from PLMs dedicated to the target language to convey richer semantic features that the source model fails to capture during the transfer. Based on this approach, we propose a new cross-lingual transfer method that recycles target language PLMs to elicit LLMs for better adaptation in target languages.

\section{SALT}
We design SALT based on the assumption that embeddings from target language-specific PLMs, mainly pre-trained on a target language corpus, may have richer semantic information than those from LLMs trained in a multilingual context with imbalanced language consideration. We explore the way of transferring embeddings from the PLM to the source LLM and aim to enhance generalization on various tasks within the target language. We show the summary of SALT framework in Figure~\ref{fig:salt} and describe the process step by step as follows:

\paragraph{Step 0: Objective} 
Given source LLM~(vocabulary embedding $\boldsymbol{E}_{s}$) with its vocabulary $V_s$ and target-language specific PLM~(vocabulary embedding  $\boldsymbol{E}_{t}$) with its vocabulary $V_t$, we want to make newly initialized target language embedding for $V_t$ to replace source LLM's embedding while minimizing contextual misalignment and maintaining semantically rich information inherent in $\boldsymbol{E}_{t}$.

\paragraph{Step 1: Subword Embedding Extraction} 
For each $V_s$ and $V_t$, we use external static embeddings from fastText~\cite{bojanowski2017enriching}, which are trained on the target languages, to extract auxiliary embeddings. We select fastText trained in multiple languages, as it can provide embeddings of various words using combinations of multiple subwords. For rare tokens that do not exist in fastText or cannot be formed via subword combinations, we initialize them from a normal distribution with mean and standard deviation from $\boldsymbol{E}_{s}$.

\paragraph{Step 2: Estimating Similarity between Shared and Non-shared Vocabulary} 
As a next step, we identify the shared vocabulary set between the $V_s$ and $V_t$. We copy the shared vocabulary embedding from $\boldsymbol{E}_{s}$. We provide further details about overlapping in Appendix~\ref{appen:salt_detail}. We assume that overlapping tokens are not specific to the target language and have likely been sufficiently learned during the source model's training stage, which is also done by previous works~\cite{dobler2023focus,liu2024ofa}. Hence, our focus is on the non-shared vocabulary set.

The most challenging task is to transfer the remaining non-shared vocabulary embeddings from $\boldsymbol{E}_{t}$ into the source model's space while preserving semantic components. We find semantically similar tokens among shared vocabulary, $V_{shared} = V_s \cap V_t$, for each non-overlapping token using extracted auxiliary embeddings. We calculate cosine similarity scores to identify the semantically nearest token in $V_{shared}$ for each $v_{t_{i}}$. We denote similarity score as in Equation~\ref{eq:sim1}, where $f$ is a fastText and $v_{o} \in V_{shared}$:

\begin{equation} \label{eq:sim1}
sim(v_{t_{i}}, v_{o}) = \frac{f(v_{t_{i}}) \cdot f(v_{o})}{\|f(v_{t_{i}})\| \|f(v_{o})\|}
\end{equation}

Then we can get semantic similarity set $C_{t_{i}}$ for each $v_{t_{i}} \notin V_{\text{shared}}$:

\begin{equation}\label{eq:simset}
C_{t_{i}} = \{sim(v_{t_{i}},v_{o}) \,|\, \forall v_{o} \in V_{shared}\}
\end{equation}

\paragraph{Step 3: Building Nearest Vocabulary Set} 
From $C_{t_{i}}$, we identify the top $k$ nearest tokens. To determine the criteria of $k$, we use Sparsemax~\cite{martins2016softmax}. Sparsemax is a variant of the softmax function that eliminates less relevant elements by assigning them values of zero, which has been adopted in prior studies~\cite{tran2020english,dobler2023focus}. 

Sparsemax on $C_{t_{i}}$ assigns weights to each paired token's similarity score. Most tokens assigned zero according to the similarity distribution in $C_{t_{i}}$ are excluded. Through this process, we extract dynamic top $k$ nearest vocabulary subsets to build a linear regression that transcribes non-shared tokens.

\paragraph{Step 4: Linear Least Square Transform}
Our primary goal is to transform non-shared vocabulary embeddings into the source model space by leveraging semantically similar shared tokens' paired embeddings. We employ the cheap and fast Least Squares Transform approach. To this end, we stack the source and target embeddings of the selected nearest tokens, which are $\boldsymbol{E}^\prime_{t_{i}}$ and $\boldsymbol{E}^\prime_{s_{i}}$ respectively. Note that selected tokens have embeddings from $\boldsymbol{E}_{s}$ and $\boldsymbol{E}_{t}$ since we only extract the nearest tokens from $V_{shared}$.

We then find the transformation matrix $X_{t_{i}}$ for a non-shared token $v_{t_{i}}$ by using $\boldsymbol{E}^\prime_{s_{i}}$ as a gold label for $\boldsymbol{E}^\prime_{t_{i}}$:

\begin{equation}\label{eq:lstsq}
\argmin_{X \in \mathbb{R}^{h_{t} \times h_{s}}} \| \boldsymbol{E}^\prime_{t_{i}}X_{t_{i}} - \boldsymbol{E}^\prime_{s_{i}}\|
\end{equation}
where $h_{s}$ and $h_{t}$ are the dimension of source and target model. The solution for each $X_{t_{i}}$ can be derived as $X_{t_{i}} = \boldsymbol{E}^{\prime+}_{t_{i}} \cdot \boldsymbol{E}^\prime_{s_{i}}$ where $\boldsymbol{E}^{\prime+}_{t_{i}}$ is a pseudo-inverse of $\boldsymbol{E}^\prime_{t_{i}}$~\cite{peters1970least}. 

Using $X_{t_{i}}$, we project each non-shared vocabulary embedding from $\boldsymbol{E}_{t}$ into $\boldsymbol{E}_{s}$ space. As we consider the similarity between tokens in fitting individual least square matrix for each target token, this semantic-aware approach enables PLM's embedding to maintain the richness of representation in the target language during the transfer and facilitates better alignment with the source model.

\paragraph{Step 5: Target Language Adaptation}
As a final step, we perform additional training with an unlabeled target language corpus, called Language Adaptive Continual Pre-training, as a final step to align the weights between transferred embedding and source model layers in the target language. This stage is crucial after the initialization or transformation of the embedding since the new embedding lacks alignment with the upper layers. Thus, this stage is considered essential as a post-transfer task~\cite{chau-etal-2020-parsing,dobler2023focus,liu2024ofa,mundra2024empirical}.

\section{Experimental Setup}

\subsection{Models}
\paragraph{Source Models} For source models, we employ the models that consider multilingual tokens in their vocabulary and have a large size of vocabulary. We employ Gemma-2b~\cite{gemmateam2024gemma} and XGLM-1.7b~\cite{lin-etal-2022-shot}, which both have vocab sizes 256k. They include vocabulary for multiple languages and consider the multilingual environment.

\paragraph{Target Models} For the target model, we select PLMs with published references of various architectures for each language to ensure reliability, focusing exclusively on base-level models. By considering these widely used standard PLMs, we ensure the generalizability of our proposed method. To be specific, we use PLMs with BERT~\cite{devlin-etal-2019-bert}, GPT~\cite{brown2020language}, and T5~\cite{raffel2020exploring} architectures for each language to verify the scalability. Unless otherwise specified, our experiments use decoder-based PLMs~(GPT) as the source model. Selected target models are provided in Appendix~\ref{appen:target}.

\subsection{Baseline}
To evaluate SALT, we compare ours against multiple strong baselines which are widely adopted for language transfer.

\paragraph{FOCUS} We establish FOCUS~\cite{dobler2023focus} as our strong baseline due to its effectiveness and success in language transfer. FOCUS trains a new tokenizer for the target language and builds target language-specific embeddings by weighted averaging the source model's embedding weights through semantic mapping with auxiliary embeddings. Given its superior performance compared to Wechsel~\cite{minixhofer2022wechsel} and the standard random initialization commonly used in cross-lingual transfer tasks, we consider FOCUS a reliable baseline for our experiments. 

\paragraph{OFA} The One For All~(OFA) Framework~\cite{liu2024ofa} uses multilingual static word vectors to inject alignment knowledge into the new subword embeddings. With the embedding factorization approach via Singular Value Decomposition, OFA initializes the embedding by leveraging a weighted average of the source embedding.

\paragraph{Multivariate} For a robust baseline, we copy the original source embeddings for the shared vocabulary like other methods and sample the non-shared vocabulary embeddings from a multivariate Gaussian distribution whose mean and variance come from the original source embedding. The only difference is in how we initialize the non-shared vocabulary.

To ensure a fair comparison, we adopt the same fastText~\footnote{\url{https://fasttext.cc/docs/en/crawl-vectors.html}} as an external static embedding for each language in OFA and FOCUS. Across all methods, we use the same tokenizer as the target model's tokenizer.

\subsection{Language Adaptive Continual Pre-training}
Following the source models' pre-training objective, we use Causal Language Modeling~(CLM) on an unlabeled corpus for language adaptation. We evaluate all benchmarks after the training and report changes in training loss over steps to evaluate convergence speed and verify the generalizability across target languages.

We use the Wikipedia dump~\cite{wikidump}, version dated 20231101, extracted from Wikipedia for training. We employ NLTK sentence tokenizer to segment the corpus into individual sentences, extracting 8M sentences for each language. Subsequently, the data is split sequentially in a 9:1 ratio into training and validation sets for model training. Hyper-parameters and training details are provided in Appendix~\ref{appen:hyper}.

\subsection{Evaluation}
In our experiments, we perform the transfer on three languages with distinct language families: German~(de), Arabic~(ar), and Vietnamese~(vi), due to the availability of diverse architectures of open-source PLMs.

\paragraph{Knowledge-based Benchmark}
We evaluate the alignment between the inherent knowledge in the model’s upper layers and the newly initialized embedding using three standard knowledge-based benchmarks: ARC~\cite{clark2018thinksolvedquestionanswering}, TruthfulQA~\cite{lin-etal-2022-truthfulqa}, and HellaSwag~\cite{zellers-etal-2019-hellaswag}. Due to limited support for non-English languages, we use the multilingual version developed by~\citet{lai2023okapiinstructiontunedlargelanguage}, which includes translations in several languages. All three benchmarks follow a multiple-choice question-answering~(MCQA) format, and we evaluate the model by measuring the log-likelihood of each answer candidate and selecting the most probable one. Accordingly, we measure accuracy for HellaSwag and normalized accuracy for ARC and TruthfulQA.

\paragraph{Machine Reading Comprehension}
We use the Machine Reading Comprehension~(MRC) task to determine whether the transferred model can accurately understand contextual meaning and implications. We assess the ability to integrate complex information to generate proper answers to questions based on a given context. We employ the MLQA~\cite{lewis2020mlqa}, XQuAD~\cite{artetxe-etal-2020-cross}, and Belebele~\cite{bandarkar-etal-2024-Belebele} datasets, each containing a context and questions. We report Exact Match~(EM) and F1-score for MLQA and XQuAD, which require the model to generate answers. For Belebele, which has MCQA configuration, we measure accuracy similar to knowledge benchmark evaluation. All models are evaluated in a three-shot setting.

\begin{table*}
\centering
\renewcommand{\arraystretch}{1.1}
\resizebox{\textwidth}{!}{
\begin{tabular}{clcccccccccccc}
\toprule
\multirow{2}{*}{\textbf{Source Model}} & \multirow{2}{*}{\textbf{Method}} & \multicolumn{4}{c}{\textbf{German}} & \multicolumn{4}{c}{\textbf{Arabic}} & \multicolumn{4}{c}{\textbf{Vietnamese}} \\
\cmidrule(lr){3-6}
\cmidrule(lr){7-10}
\cmidrule(lr){11-14}
&  & \textbf{ARC} & \textbf{HS} & \textbf{TQA} & \textbf{Avg} & \textbf{ARC} & \textbf{HS} & \textbf{TQA} & \textbf{Avg} & \textbf{ARC} & \textbf{HS} & \textbf{TQA} & \textbf{Avg} \\
\toprule

\multirow{4}{*}{XGLM} 
& Multivariate & 24.72 & 29.66 & 26.27 & 26.88 & \textbf{26.69} & 26.37 & 26.78 & 26.61 & 25.64 & 32.61 & 28.41 & 28.89 \\
& FOCUS & 23.95 & 29.89 & 26.90 & 26.91 & 25.15 & 26.74 & \textbf{28.72} & 26.87 & 24.87 & 33.13 & 28.54 & 28.85 \\
& OFA & 24.12 & 30.47 & 26.90 & 27.16 & 24.55 & 28.36 & 27.94 & 26.95 & 24.87 & 32.50 & 28.54 & 28.64 \\
& SALT (Ours) & \textbf{25.49} & \textbf{31.73} & \textbf{27.28} & \textbf{28.17} & 25.24 & \textbf{29.27} & 28.07 & \textbf{27.53} & \textbf{25.73} & \textbf{33.69} & \textbf{28.54} & \textbf{29.32} \\

\midrule

\multirow{4}{*}{Gemma} 
& Multivariate & 29.43 & 39.26 & 24.37 & 31.02 & 26.43 & 31.68 & 28.98 & 29.03 & 27.61 & 42.60 & 29.43 & 33.21 \\
& FOCUS & 26.95 & 38.27 & 24.24 & 29.82 & 25.66 & 29.86 & 28.98 & 28.17 & 27.86 & 42.15 & 28.28 & 32.76 \\
& OFA & 28.74 & 38.74 & 24.62 & 30.70 & 26.52 & 33.82 & 28.07 & 29.47 & 26.58 & 41.80 & 29.17 & 32.52 \\
& SALT (Ours) & \textbf{30.80} & \textbf{42.58} & \textbf{24.75} & \textbf{32.71} & \textbf{27.03} & \textbf{35.90} & \textbf{28.98} & \textbf{30.64} & \textbf{30.77} & \textbf{43.77} & \textbf{29.43} & \textbf{34.66} \\

\bottomrule
\end{tabular}}
\caption{Performance on knowledge-based benchmarks. HS and TQA stand for HellaSwag and TruthfulQA. We \textbf{bold} the best result in each section under the same condition.}
\label{tab:knowledge}
\end{table*}

\begin{figure*}[hbt!]
\centering 
\includegraphics[width=0.98\linewidth]{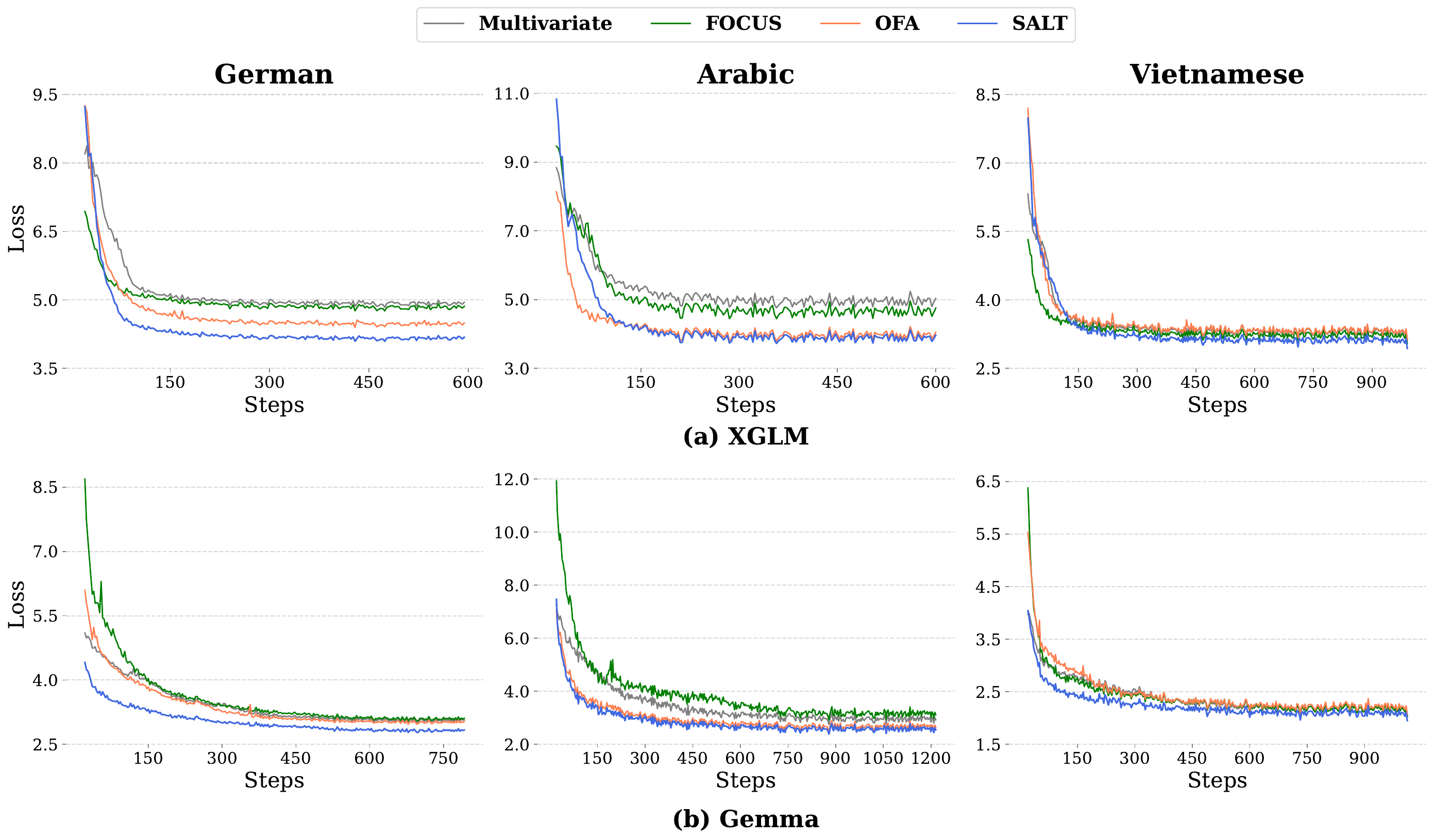}
\caption{Causal Language Modeling~(CLM) loss of language adaptive continual pre-training. The x-axis represents the progression of training steps, while the y-axis denotes the training loss. The first data point is logged at 20 steps.}
\label{fig:loss} 
\end{figure*}

\paragraph{Cross-lingual Question Answering}
We also evaluate the model’s linguistic ability in a cross-lingual setting where English and the target language coexist. Among the datasets employed in our MRC evaluation, MLQA encompasses instances where the context and question are in different languages. We report the performance of transferred models under both English-target language and target language-English configurations. Through comparisons with other approaches, we assess the inner alignment between English and the target language and understanding ability.

\section{Results}
We present downstream task performance for knowledge-based benchmarks in Section~\ref{sec:knowledge-bench}. In Section~\ref{sec:loss_result}, we show the loss curves during language adaptive continual pre-training with each transfer method. We also report MRC performance in a monolingual and cross-lingual setup in Section~\ref{sec:MRC}. Additionally, we verify the extensibility of our approach by examining the performance when using various types of target language-specific PLMs beyond the decoder type in Section~\ref{sec:extension}.

\subsection{Inherent Knowledge Alignments}
\label{sec:knowledge-bench}
Table~\ref{tab:knowledge} shows the effectiveness of SALT in aligning inherent knowledge in the source model. Regardless of the source model, SALT consistently outperforms all baselines across tasks and languages on average, delivering significant performance gains.

\begin{table*}
\centering
\renewcommand{\arraystretch}{1.1}
\resizebox{\textwidth}{!}{
\begin{tabular}{clccccccccc}
\toprule
\multirow{2}{*}{\textbf{Lang}} & \multirow{2}{*}{\textbf{Method}} & \multicolumn{3}{c}{\textbf{MLQA}} & \multicolumn{3}{c}{\textbf{XQuAD}} & \multicolumn{3}{c}{\textbf{Belebele}} \\
\cmidrule(lr){3-5}
\cmidrule(lr){6-8}
\cmidrule(lr){9-11}
& & \textbf{German} & \textbf{Arabic} & \textbf{Vietnamese} & \textbf{German} & \textbf{Arabic} & \textbf{Vietnamese} & \textbf{German} & \textbf{Arabic} & \textbf{Vietnamese} \\
\toprule

\multirow{4}{*}{Target} 
& Multivariate & 19.70 / 31.68 & 10.35 / 25.37 & 23.44 / 42.16 & 22.23 / 34.62 & 21.67 / 31.50 & 31.01 / 51.74 & 33.22 & \textbf{27.89} & 26.22 \\
& FOCUS & 25.66 / 37.97 & 12.63 / 25.63 & 27.17 / 44.77 & 27.21 / 39.00 & 17.09 / 25.01 & 35.71 / 54.65 & 31.00 & 26.67 & 31.33 \\
& OFA & 23.22 / 37.05 & 16.49 / 33.52 & 25.06 / 45.14 & 26.52 / 38.99 & 22.27 / 32.58 & 35.88 / 55.07 & 28.67 & 27.67 & 31.56 \\
& SALT (Ours) & \textbf{28.07 / 41.47} & \textbf{18.33 / 36.36} & \textbf{29.70 / 49.63} & \textbf{31.28 / 44.41} & \textbf{27.07 / 39.19} & \textbf{36.30 / 56.90} & \textbf{34.11} & 27.67 & \textbf{35.78} \\
\midrule

\multirow{4}{*}{English} 
& Multivariate & 30.22 / 46.12 & 36.52 / 51.42 & 33.38 / 47.50 & 33.70 / 48.25 & 40.17 / 53.51 & 36.05 / 49.27 & 32.00 & \textbf{32.56} & 32.33 \\
& FOCUS & 35.66 / 50.26 & 38.00 / 52.57 & 36.82 / 50.56 & 40.34 / 52.43 & 40.25 / 53.45 & 41.34 / 53.06 & 31.00 & 28.44 & 32.00 \\
& OFA & 31.94 / 46.94 & 35.28 / 49.75 & 35.32 / 50.46 & 35.63 / 49.97 & 37.98 / 49.80 & 40.25 / 54.34 & 28.67 & 26.33 & 32.56 \\
& SALT (Ours) & \textbf{40.02 / 55.01} & \textbf{40.24 / 54.96} & \textbf{43.64 / 58.54} & \textbf{43.78 / 57.25} & \textbf{42.10 / 54.35} & \textbf{48.24 / 61.85} & \textbf{33.67} & 30.67 & \textbf{35.22} \\

\bottomrule
\end{tabular}}
\caption{Results of Gemma initialized with SALT and baselines in monolingual MRC tasks. Exact Match / F1-score for MLQA and XQuAD, and accuracy for Belbele are reported. The ``Lang'' column indicates the result under the target language or English.}
\label{tab:mrc}
\end{table*}

\begin{table*}
\centering
\renewcommand{\arraystretch}{1.2}
\resizebox{\textwidth}{!}{
\begin{tabular}{lcccccccc}
\toprule
\multirow{2}{*}{\textbf{Method}} & \multicolumn{4}{c}{\textbf{English - Target}} & \multicolumn{4}{c}{\textbf{Target - English}} \\
\cmidrule(lr){2-5}
\cmidrule(lr){6-9}
& \textbf{German} & \textbf{Arabic} & \textbf{Vietnamese} & \textbf{Avg} & \textbf{German} & \textbf{Arabic} & \textbf{Vietnamese} & \textbf{Avg} \\
\toprule
Multivariate & 22.23 / 34.62 & 21.67 / 31.50 & 22.58 / 33.06 & 22.16 / 33.06 & 18.51 / 29.08 & 11.68 / 23.89 & 22.57 / 40.17 & 17.59 / 31.05 \\
FOCUS & 27.21 / 39.00 & 17.09 / 25.01 & 25.02 / 33.54 & 23.11 / 32.52 & 24.51 / 35.66 & 14.26 / 25.69 & 26.99 / 42.76 & 21.92 / 34.70 \\
OFA & 26.52 / 38.99 & 22.17 / 32.58 & 25.30 / 36.41 & 24.66 / 35.99 & 19.46 / 30.96 & 16.16 / 29.44 & 24.53 / 41.38 & 20.05 / 33.93 \\
SALT (Ours) & \textbf{31.28 / 44.41} & \textbf{27.07 / 39.19} & \textbf{29.30 / 40.59} & \textbf{29.22 / 41.40} & \textbf{26.59 / 38.79} & \textbf{17.11 / 30.92} & \textbf{29.17 / 46.15} & \textbf{24.29 / 38.62} \\
\bottomrule
\end{tabular}}
\caption{Results of MLQA in cross-lingual setup in English with the target language. On the top of the Table, the first language denotes the context's language, while the second language denotes the question's language.}
\label{tab:crossmrc}
\end{table*}


The major difference is evident in HellaSwag. In Gemma, SALT obtains 42.58 points on German HellaSwag, exceeding the second-best multivariate initialization by more than three points. In HellaSwag, models are required to select a sentence that logically follows a given context. This shows that SALT is superior in capturing the linguistic context and logical flow of the target language. The gains mainly arise from whether the models utilize PLM embedding trained on the target language. Since PLM embedding has already experienced the alignment with target language knowledge during pre-training, SALT further promotes alignment between the embedding layer and knowledge in the upper layers, which is also observed in Figure~\ref{fig:loss}.

In contrast, FOCUS and OFA often yield similar or lower scores than the multivariate, which lacks semantic considerations for target language embeddings. Although both FOCUS and OFA incorporate target language semantics, the consideration stems from a shallow level of representation of the target language may harm alignment. This finding stresses the need for new transfer methods that differ from the weighted average of the source model's embedding.

\subsection{Generalization from Continual Pre-training}
\label{sec:loss_result}
To analyze how different initialization methods influence continual pre-training, we visualize the training loss of models initialized with SALT and all baselines. In Figure~\ref{fig:loss}, models initialized with SALT generally converge faster and yield the lowest training loss at the last step in every scenario~(Appendix~\ref{appen:loss}). In settings with limited computing resources, SALT may be even more beneficial. Even for Gemma, it maintains the lowest loss from the beginning to the end. Unexpectedly, FOCUS consistently exhibits higher loss in German and Arabic compared to SALT and OFA. This demonstrates that initializing embeddings from target language-specific PLMs while considering semantic features can be superior during the adaptation period, underscoring the benefits of adopting them for language transfer.

\begin{table*}
\centering
\renewcommand{\arraystretch}{1.1}
\resizebox{\textwidth}{!}{
\begin{tabular}{clcccccccccccc}
\toprule
\multirow{2}{*}{\textbf{Source Model}} & \multirow{2}{*}{\textbf{Target Model}} & \multicolumn{4}{c}{\textbf{German}} & \multicolumn{4}{c}{\textbf{Arabic}} & \multicolumn{4}{c}{\textbf{Vietnamese}} \\
\cmidrule(lr){3-6}
\cmidrule(lr){7-10}
\cmidrule(lr){11-14}
&  & \textbf{ARC} & \textbf{HS} & \textbf{TQA} & \textbf{Avg} & \textbf{ARC} & \textbf{HS} & \textbf{TQA} & \textbf{Avg} & \textbf{ARC} & \textbf{HS} & \textbf{TQA} & \textbf{Avg} \\
\toprule

\multirow{3}{*}{XGLM} 
& BERT & 25.15 & 32.13 & \textbf{28.05} & \textbf{28.44} & 23.78 & 27.95 & 28.05 & 26.59 & 25.13 & 31.83 & \textbf{30.45} & 29.14 \\
& GPT & \textbf{25.49} & 31.73 & 27.28 & 28.17 & \textbf{25.24} & 29.27 & \textbf{28.07} & \textbf{27.53} & \textbf{25.73} & \textbf{33.69} & 28.54 & \textbf{29.32} \\
& T5 & 24.47 & \textbf{32.57} & 27.92 & 28.32 & 24.38 & \textbf{29.48} & 27.92 & 27.26 & 25.73 & 33.67 & 28.41 & 29.27 \\

\midrule

\multirow{3}{*}{Gemma} 
& BERT & 30.37 & 42.70 & 25.00 & 32.69 & 28.57 & 34.29 & 25.00 & 29.29 & 27.35 & 37.43 & 22.29 & 29.02 \\
& GPT & \textbf{30.80} & 42.58 & 24.75 & 32.71 & 27.03 & 35.90 & \textbf{28.98} & \textbf{30.64} & \textbf{30.77} & 43.77 & \textbf{29.43} & \textbf{34.66} \\
& T5 & 30.71 & \textbf{43.13} &\textbf{25.51}& \textbf{33.12} & \textbf{28.66} & \textbf{36.52} & 25.51 & 30.23 & 30.17 & \textbf{43.82} & 27.77 & 33.92 \\

\bottomrule
\end{tabular}}
\caption{Performance across different target model architectures in knowledge benchmarks.}
\label{tab:bert_t5}
\end{table*}

\subsection{Effects In Language Understanding}
\label{sec:MRC}
Table~\ref{tab:mrc} and Table~\ref{tab:crossmrc} show the results in two different setups: the target language setup, which includes English, and a cross-lingual setup.

\paragraph{Target Language MRC}
At the top of Table~\ref{tab:mrc}, SALT shows its superiority in MRC tasks in the target language. The gap from the baselines is especially large in MLQA and XQuAD, both of which adopt generation-based evaluations. Unlike other baselines that rely on embedding from the source model, SALT may benefit from the richer semantic information specialized in the target language inherited from PLM's embedding. This approach allows transferred models to capture contextual cues better and generate correct answers by leveraging necessary linguistic information more accurately in language comprehension tasks.

However, despite achieving performance on par with FOCUS and OFA in previous results, multivariate initialization records the lowest scores on MLQA and XQuAD. An initialization method ignoring the target language’s semantic information struggles to capture the deeper inferences and contextual cues needed for high-level reasoning based on external context.

\paragraph{Influence in Cross-lingual Understanding}
We find that SALT is more effective in understanding the mutual context in a cross-lingual environment. In Table~\ref{tab:crossmrc}, SALT outperforms other approaches when English and the target language coexist within the same sample. In the English context with the target language question setup, SALT obtains 29.22 EM and 41.40 F1-score on average, surpassing the second-best baseline OFA by about 4 and 5 points, respectively—a 13\% improvement in F1-score. We observe similar trends when the languages of the context and question are reversed. SALT outperforms FOCUS by approximately 2.4 points in EM and 4 points in F1-score.

In general, most PLMs mainly focusing on the target language often consider both English and the target language in the training stage, creating a shared space for both. By projecting this fused space to a new embedding, SALT successfully blends the source model's English ability with newly initialized embeddings for the target language, leading to notable benefits in cross-lingual understanding.

Likewise, it is evidenced by the result of the English evaluation at the bottom of Table~\ref{tab:mrc}. Our direct evaluation in English confirms that SALT preserves English proficiency better than other transfer methods, even though each model is further trained only on the target language. This highlights the greater effectiveness of SALT in cross-lingual settings and its utility in general situations that require both the target language and English.

\subsection{Extension to Target Models with Different Architectures}
\label{sec:extension}
To demonstrate the scalability of SALT, we compare performance using target models whose architectures differ from the source model~(e.g., BERT and T5). Their embedding formations differ based on learning objectives. We find that SALT can be a valid transfer method for target models with different structures from the source model.

As illustrated in Table~\ref{tab:bert_t5}, the decoder-based target model mostly achieves the best average results. However, employing PLMs with different structures as the target model performs better in some cases. For German, using T5 generally outperforms GPT in both source models. In addition, for German and Arabic, BERT’s performance gap with GPT remains within about one point on average.

We also report the evaluation loss according to various types of target models in SALT in Table~\ref{tab:eval_loss_arch}. In some cases, BERT achieves lower evaluation loss compared to GPT. Although learning patterns may vary depending on the underlying target model, these observations suggest that other types of models can be integrated into decoder-only LLMs from a transfer standpoint. Given that various kinds of PLMs exist, SALT can be a flexible method that is not restricted to decoder-type target models and can be extended to various architectures.
\begin{table}[t]
\centering
\renewcommand{\arraystretch}{1.15}
\resizebox{\linewidth}{!}{
\begin{tabular}{ll*{9}{c}}
\toprule
\multirow{2}{*}{\textbf{Source}} & \multirow{2}{*}{\textbf{Target}} 
& \multicolumn{3}{c}{\textbf{German}} 
& \multicolumn{3}{c}{\textbf{Arabic}} 
& \multicolumn{3}{c}{\textbf{Vietnamese}} \\
\cmidrule(lr){3-5}\cmidrule(lr){6-8}\cmidrule(lr){9-11}
 & & \textbf{30\%} & \textbf{60\%} & \textbf{90\%} 
   & \textbf{30\%} & \textbf{60\%} & \textbf{90\%} 
   & \textbf{30\%} & \textbf{60\%} & \textbf{90\%} \\
\midrule
\multirow{3}{*}{XGLM}
 & BERT & 3.89 & 3.82 & 3.80 & 3.76 & 3.25 & 3.24 & 3.49 & 3.40 & 3.40 \\
 & GPT  & 4.08 & 4.00 & 3.99 & 3.83 & 3.68 & 3.67 & 3.25 & 3.18 & 3.17 \\
 & T5   & 3.77 & 3.71 & 3.70 & 4.27 & 4.14 & 4.13 & 3.07 & 3.02 & 3.01 \\
\midrule
\multirow{3}{*}{Gemma}
 & BERT & 2.81 & 2.77 & 2.73 & 2.39 & 2.24 & 2.22 & 2.50 & 2.45 & 2.37 \\
 & GPT  & 3.09 & 2.88 & 2.81 & 2.76 & 2.54 & 2.51 & 2.37 & 2.32 & 2.24 \\
 & T5   & 3.07 & 2.90 & 2.85 & 3.15 & 2.91 & 2.86 & 2.32 & 2.29 & 2.22 \\
\bottomrule
\end{tabular}
}
\caption{Evaluation loss at various percentages of total steps for each target model in SALT.}
\label{tab:eval_loss_arch}
\end{table}

\section{Conclusion}
We propose SALT, a novel cross-lingual transfer method that recycles the embeddings of PLMs for transferring from English-centric LLMs to target language-specific LLMs. By deriving a unique linear regression line that takes into account the semantic similarity for each non-overlapping vocabulary embedding, we map the rich semantic features in PLM's embedding onto the LLM's embedding. A series of experiments reveals that SALT benefits from the rich expressiveness of PLM's embedding in the target language. Also, SALT can promote better initialization and faster convergence for language adaptation, enhancing generalization to the target language. Notably, among existing initialization methods, SALT preserves English capabilities more effectively and demonstrates greater superiority in cross-lingual setups. We also investigate expandability by experimenting with multiple types of target models, indicating that SALT can be applied to various PLM architectures. For future work, we plan to explore methods that facilitate the transfer of target language-specific knowledge embedded within model layers.

\section*{Limitations}
Our methodology is based on target language-specialized PLMs, making their presence essential. Our selection of languages in this paper is based on this consideration. For our experiments in diverse types of PLMs, low-resource languages where these PLMs are unavailable are out of our scope. Given that SALT benefits from target language PLMs’ rich representation ability, SALT is likely to be an even more effective method in low-resource languages if PLMs exist.

In the evaluation, we did not investigate the impact of instruction tuning. Our focus lies in leveraging foundational models for transfer and evaluating the benefits derived from PLM embeddings and the linguistic ability of the transferred model, since our approach is only related to the initialization stage. Our transfer method provides a robust starting point for embeddings, laying the groundwork for future instruction tuning to build models that perform better in the target language.

Also, we evaluate SALT only for models with large vocabulary sizes accounting for a relatively heavy portion of embedding in the model’s total parameters. Given that one of the objectives of cross-lingual transfer is mitigating overload caused by irrelevant tokens to the target language and pursuing effectiveness, this aligns with the goal. In principle, SALT can also be applied to larger models~(>7b), and language-specific LLMs can be used for target models instead of PLMs. However, our focus is on the usability of the previous PLMs and we would call for exploration in this direction.

\section*{Ethical Statement}
In this study, we utilized publicly available LLMs and PLMs. These models were chosen from open sources that meet ethical standards. Additionally, the static embeddings used for cross-lingual transfer and the data used in continual pre-training were sourced from publicly accessible resources and applied in accordance with appropriate licenses to align with the research objectives. Furthermore, we are aware of the potential biases that might be present in the foundation model, which could be inadvertently transferred to target languages during the language transfer process, and such biases may lead to differential performance or unintended outcomes across various languages and societal groups.

\section*{Acknowledgments}
This research was supported by Basic Science Research Program through the National Research Foundation of Korea~(NRF) funded by the Ministry of Education~(NRF-2021R1A6A1A03045425). This work was supported by Institute for Information \& communications Technology Promotion~(IITP) grant funded by the Korea government~(MSIT) 
(RS-2024-00398115, Research on the reliability and coherence of outcomes produced by Generative AI). This work was supported by Institute of Information \& communications Technology Planning \& Evaluation~(IITP) under the Leading Generative AI Human Resources Development~(IITP-2025-R2408111) grant funded by the Korea government~(MSIT). This work was supported by Institute of Information \& communications Technology Planning \& Evaluation~(IITP) under the artificial intelligence star fellowship support program to nurture the best talents~(IITP-2025-RS-2025-02304828) grant funded by the Korea government~(MSIT). 

\bibliography{custom}

\appendix

\section{Further Details of SALT}
\label{appen:salt_detail}
\paragraph{Token Overlap}
SALT determines token overlap to transfer the target model’s non-overlapping vocabulary into the source model’s space. In this paper, we consider various target models with distinct tokenizers; GPT uses Byte-Level BPE, BERT uses WordPiece, and T5 uses SentencePiece. The main difference between tokenizers is how they handle whitespace tokens and non-ASCII character tokens. For example, HuggingFace’s Byte-Level BPE implementation prefixes whitespace tokens with `Ġ', SentencePiece uses `\_', and BERT’s WordPiece uses `\#\#'. To avoid these discrepancies, we remove all white spaces and only consider meaningful characters in each token, which is similar to `fuzzy search' in FOCUS~\cite{dobler2023focus}.

\paragraph{Vocabulary Coverage}
We report the number of overlapping tokens for each language and source model, as well as the number of tokens transferred from PLMs for initialization in Table~\ref{tab:coverage}. Coverage refers to the proportion of the target PLM’s vocabulary contained in the source vocabulary. Coverage varies according to the source model and language. Vietnamese generally shows the highest overlap, while Arabic exhibits the lowest. 

Insufficient overlap potentially hinders the formation of a linear regression line conveying the capabilities of target-language specialized PLMs. Consequently, comparatively lower performance in Arabic than in other languages in our experiments may stem from the low coverage rate of overlapping tokens.

\paragraph{Implementation}
We apply the same mechanism to both the embedding layer and the LM head layer. Accordingly, both the embedding layer and the head layer are initialized with the vocabulary of the target language model. If the LM embedding layer has dimensions of (vocab\_size × hidden\_size), then the head layer adopts dimensions of (hidden\_size × vocab\_size), which is the transpose of the LM embedding matrix. However, this may vary depending on whether the model uses tied embedding weights. In our case, we used a model where the LM head and the embedding layer share the same weights; therefore, we initialized the new embedding matrix and then assigned its transpose to the LM head. If the weights are not shared, it would be necessary to apply SALT to the LM head weight matrix separately.
\begin{table}[t]
    \centering
    \renewcommand{\arraystretch}{1.1}
    \resizebox{\linewidth}{!}{
    \begin{tabular}{ccccc}
            \toprule
            \textbf{Lang} &\textbf{Model} & \textbf{Copied} & \textbf{Initialized} & \textbf{Coverage} (\%) \\
            \toprule
            \multirow{2}{*}{De} & XGLM & 25,254 & 24,853 & 50.2 \\
            & Gemma & 29,369 & 20,852 & 58.4 \\
            \midrule        
            \multirow{2}{*}{Ar} & XGLM & 10,758 & 53,070 & 21.4 \\
            & Gemma & 7,851 & 56,006 & 15.6 \\
            \midrule        
            \multirow{2}{*}{Vi} & XGLM & 30,992 & 18,736 & 61.7 \\
            & Gemma & 34,114 & 15,934 & 67.9 \\
            \bottomrule        
    \end{tabular}}
    \caption{The number of tokens being initialized or copied from the original embeddings.}
    \label{tab:coverage}
\end{table}

\section{Selected Target Models}
\label{appen:target}
We list the target PLMs used in our experiments in Table~\ref{tab:appen_models}. We prefer PLMs with published references to ensure reliability. If no relevant paper exists, we select widely used pre-trained PLMs that have not been fine-tuned for downstream tasks, which have a large number of downloads on the HuggingFace Model Hub.
\begin{table}[hbt!]
\centering
\resizebox{0.95\linewidth}{!}{
\begin{tabular}{cp{9cm}}
\toprule
\textbf{Lang} & \multicolumn{1}{c}{\textbf{Target Model}} \\
\midrule

\multirow{6}{*}{De} & \texttt{- deepset/gbert-base} \newline \cite{german-bert}\\
 & \texttt{- dbmdz/german-gpt2} \newline \cite{german-gpt} \\
 & \texttt{- GermanT5/t5-efficient-gc4-german-base-nl36} \newline \cite{german-t5} \\
\midrule
 
\multirow{6}{*}{Ar} & \texttt{- aubmindlab/bert-base-arabertv2} \newline \cite{antoun2020arabert} \\
 & \texttt{- aubmindlab/aragpt2-medium} \newline \cite{arabic-gpt} \\
 & \texttt{- UBC-NLP/AraT5v2-base-1024} \newline \cite{elmadany-etal-2023-octopus} \\
\midrule
 
\multirow{6}{*}{Vi} & \texttt{- vinai/phobert-base-v2} \newline \cite{phobert} \\
 & \texttt{- NlpHUST/gpt2-vietnamese} \newline  \cite{viet-gpt} \\
 & \texttt{- VietAI/vit5-base} \newline  \cite{phan-etal-2022-vit5} \\
\bottomrule

\end{tabular}}
\caption{Target model~(PLMs) used for our experiments.}
\label{tab:appen_models}
\end{table}

\section{Training Details}
\label{appen:hyper}
We conduct all experiments using a uniform setup across all languages, employing two NVIDIA A100 GPUs, each with 80GB of memory, to perform language-adaptive continual pre-training. For Gemma, we set a maximum length of 1024 and utilized a batch size of 32 with gradient accumulation step to 4, resulting in a global batch size of 512~(32 x 4 x 4) and trained single epoch. The learning rate is established at 1e-5, using a cosine scheduler and a warmup ratio of 0.05. All other hyper-parameters remain consistent across the models. We adopt the Pytorch framework~\cite{paszke2019pytorch} and the Huggingface Transformers library~\cite{wolf2020transformers}. For knowledge benchmark evaluation, we utilize the Language Model Evaluation Harness framework~\footnote{\url{https://github.com/EleutherAI/lm-evaluation-harness}} for evaluation.

\section{Training loss}
\label{appen:loss}
We report the training loss at the final step in Table~\ref{tab:final_loss}. As mentioned in the paper, SALT records the lowest training loss in the last step. 
\begin{table}[hbt!]
\centering
\renewcommand{\arraystretch}{1.1}
\resizebox{\linewidth}{!}{
\begin{tabular}{clccc}
\toprule
\textbf{Model} & \textbf{Method} & \textbf{German} & \textbf{Arabic} & \textbf{Vietnamese} \\
\midrule
\multirow{4}{*}{XGLM} 
& Multivariate & 4.9465 & 5.0360 & 3.1243 \\
& FOCUS & 4.8564 & 4.7522 & 3.0472 \\
& OFA & 4.4922 & 4.0532 & 3.1390 \\
& SALT & \textbf{4.1823} & \textbf{3.9648} & \textbf{2.9373} \\

\midrule

\multirow{4}{*}{Gemma} 
& Multivariate & 3.0726 & 2.9218 & 2.0330 \\
& FOCUS & 3.1088 & 3.1154 & 2.0319 \\
& OFA & 3.0326 & 2.6474 & 2.0671 \\
& SALT & \textbf{2.8334} & \textbf{2.5489} & \textbf{1.9509} \\

\bottomrule
\end{tabular}}
\caption{The training loss at the final step.}
\label{tab:final_loss}
\end{table}

\section{Length \& Parameter Footprint Analysis}
We also report the efficiency benefits of language transfer from a practical standpoint. Table~\ref{tab:tok_len} presents a comparison of tokenized lengths for randomly extracted 100,000 samples per language from the training dataset. 

The result shows a notable reduction in tokenized length for the identical sentences when employing SALT compared to the original LLMs. For Arabic, the decrease in Gemma is about 25\%, highlighting the lack of consideration for language-specific vocabulary in the original LLM. This reduction is closely related to computational efficiency. The generalized tokenization strategy of original LLMs, which neglects language-specific traits, can lead to a considerable computational burden during training or inference.

Similarly, SALT results in a substantial decrease of parameters: 24\% in XGLM and 17\% in Gemma. The reduction is due to a decrease in embedding parameters by eliminating vocabulary that is unnecessary in the target language within LLMs. SALT reduces unnecessary loads for the target language and enables the use of a language-specific tokenization strategy that accounts for the unique characteristics of the target language derived from a considerable amount of monolingual corpus.
\begin{table}[t]
\centering
\renewcommand{\arraystretch}{1.1} 
\resizebox{\linewidth}{!}{
\begin{tabular}{cccc} 
\toprule
\multirow{2}{*}{\textbf{Model}}  & \multicolumn{3}{c}{\textbf{Language}} \\
 & \textbf{German} & \textbf{Arabic} & \textbf{Vietnamese} \\ 
\midrule
\multicolumn{4}{c}{\textbf{\textit{Avg. Tokenized Length}}} \\
\midrule
XGLM-1.7b                       & 30.95       & 44.47       & 40.54       \\ 
Gemma-2b                        & 31.62       & 51.18       & 43.33       \\ 
\multirow{2}{*}{SALT}            & \textbf{28.21}       & \textbf{38.61}        & \textbf{38.64}     \\
                                & (9\% / 11\%)$\downarrow$ &  (13\% / 25\%)$\downarrow$ & (5\% / 11\%)$\downarrow$ \\
\midrule
\multicolumn{4}{c}{\textbf{\textit{\# of parameters}}} \\
\midrule
\multirow{2}{*}{SALT}            & \textbf{1.31B / 2.08B}       & \textbf{1.34B / 2.11B}        & \textbf{1.31B / 2.08B}     \\
                                & (24\% / 17\%)$\downarrow$ &  (23\% / 16\%)$\downarrow$ & (24\% / 17\%)$\downarrow$ \\
\bottomrule
\end{tabular}}
\caption{Reduction in average tokenized length and the number of total parameters according to original LLMs and after SALT. We report the percentage reduction compared to the original LLMs with `XGLM / Gemma'.}
\label{tab:tok_len}
\end{table}

\end{document}